\documentclass[runningheads]{llncs}
\usepackage{graphicx}

\usepackage{tikz}
\usepackage{comment}
\usepackage{amsmath,amssymb} 
\usepackage{color}
\usepackage{cite}

\usepackage[accsupp]{axessibility}  

\usepackage[labelfont=bf, font=small, labelsep=period]{caption}
\DeclareMathOperator*{\argmin}{argmin}

\newcommand{\ie}{\textit{i}.\textit{e}.}
\newcommand{\eg}{\textit{e}.\textit{g}.}
\usepackage{makecell}
\usepackage{multirow}
\usepackage{booktabs}
\usepackage{pifont}

\usepackage{wrapfig}
\usepackage{orcidlink}
\usepackage{hyperref}
\hypersetup{
  colorlinks=true, 
  urlcolor=black, 
  linkcolor=black, 
  citecolor=green 
}

\begin{document}
\pagestyle{headings}
\mainmatter
\def\ECCVSubNumber{5537}  

\title{Speaker-adaptive Lip Reading with User-dependent Padding} 

\titlerunning{Speaker-adaptive Lip Reading with User-dependent Padding}
%
\author{Minsu Kim\orcidlink{0000-0002-6514-0018} \and
Hyunjun Kim\orcidlink{0000-0001-6524-8689} \and
Yong Man Ro\orcidlink{0000-0001-5306-6853}}
\authorrunning{M. Kim et al.}
%
\institute{Image and Video Systems Lab, School of Electrical Engineering, KAIST\\
\email{\{ms.k, kimhj709, ymro\}@kaist.ac.kr}
}
\maketitle

\begin{abstract}
Lip reading aims to predict speech based on lip movements alone. As it focuses on visual information to model the speech, its performance is inherently sensitive to personal lip appearances and movements. This makes the lip reading models show degraded performance when they are applied to unseen speakers due to the mismatch between training and testing conditions. Speaker adaptation technique aims to reduce this mismatch between train and test speakers, thus guiding a trained model to focus on modeling the speech content without being intervened by the speaker variations. In contrast to the efforts made in audio-based speech recognition for decades, the speaker adaptation methods have not well been studied in lip reading. In this paper, to remedy the performance degradation of lip reading model on unseen speakers, we propose a speaker-adaptive lip reading method, namely user-dependent padding. The user-dependent padding is a speaker-specific input that can participate in the visual feature extraction stage of a pre-trained lip reading model. Therefore, the lip appearances and movements information of different speakers can be considered during the visual feature encoding, adaptively for individual speakers. Moreover, the proposed method does not need 1) any additional layers, 2) to modify the learned weights of the pre-trained model, and 3) the speaker label of train data used during pre-train. It can directly adapt to unseen speakers by learning the user-dependent padding only, in a supervised or unsupervised manner. Finally, to alleviate the speaker information insufficiency in public lip reading databases, we label the speaker of a well-known audio-visual database, LRW, and design an unseen-speaker lip reading scenario named LRW-ID. The effectiveness of the proposed method is verified on sentence- and word-level lip reading, and we show it can further improve the performance of a well-trained model with large speaker variations.

\end{abstract}

\section{Introduction}

Lip reading, also known as Visual Speech Recognition (VSR), aims to predict what a person is saying based on visual information alone. It has drawn big attention with its beneficial applications, such as speech recognition under a noisy environment, extracting speech of target speaker from multi-speaker overlapped speech, and conversation with people who cannot make a voice. 
With the great development of deep learning and the availability of large-scale audio-visual databases \cite{cooke2006grid,chung2016lrw,chung2017lrs2,afouras2018lrs3,yang2019lrw1000}, many efforts have been made to improve lip reading performance. Architectural improvements of deep neural network are made by \cite{assael2016lipnet,stafylakis2017resnetlstm,petridis2018end,afouras2018deep,martinez2020mstcn}, pre-training schemes are introduced by \cite{chung2016syncnet,ma2021lira}, and coupling audio modal knowledge into lip reading is performed by \cite{zhao2020hearing,afouras2020asrisallyouneed,ren2021learningfrommaster,kim2021cromm,kim2021mmbridge,kim2022distinguishing}. 

It is widely known that speech recognition techniques, including both audio-based Automatic Speech Recognition (ASR) and lip reading, show degraded performance when they are applied on unseen speakers due to the mismatch between train and test data distributions \cite{digalakis1995unseen,assael2016lipnet,klejch2019meta}. Speaker adaptation technique aims to narrow this mismatch by fitting a trained speech recognition model to unseen test speakers to improve performances during test time.
With its practical importance, speaker adaptation has been an important research topic in ASR for decades \cite{anastasakos1997mllr,miao2014towardsSAT,miao2015SATivector,klejch2019meta,neto1995lin,xue2014fastadaptation,huang2020SAspeechsynthesis}. They attempt to optimize the speech recognition performance by transforming pre-trained models to well operate on one particular speaker or modifying the encoded features to match the pre-trained model, by using a small amount of adaptation data. For example, previous works \cite{abdel2013codecnn,abdel2013codeicassp,miao2015SATivector,xue2014fastadaptation} showed that providing speaker-specific input as hints for the input speaker to the ASR model is beneficial in adapting the trained model to an unseen speaker.

However, in contrast to the efforts in ASR, speaker adaptation methods have not been well addressed in lip reading. Since different speakers show varying lip appearances and movements, it is also important in lip reading to adaptively encode the lips of different speakers to achieve robust performance.
As lip reading handles lip movement video which is higher-dimensional than audio (\ie, composed of both spatial and temporal dimensions), encoding spatio-temporal information to be aware of the displacement of lips and their movement is important for accurate recognition. To this end, visual features are usually extracted using 2D or 3D Convolutional Neural Network (CNN) to achieve high recognition performance \cite{noda2014lipreadingconv,assael2016lipnet}, compared to the discriminative audio features that are relatively easily obtained by transforming the raw audio into Mel-Frequency Cepstral Coefficient (MFCC) or Mel-spectrogram in ASR. Due to the different characteristics of modalities and feature extraction methods, the speaker adaptation methods of ASR might be less effective when they are directly applied to lip reading. Therefore, a speaker adaptation method suitable for lip reading, which can jointly consider the spatial information of visual features during adaptation is required. One main impediment in developing speaker-adaptive lip reading is the lack of speaker information in public databases. Usually, publicly available large-scale lip reading databases \cite{chung2016lrw,yang2019lrw1000,chung2017lrs2} have no speaker information and have overlapped speakers between train and test splits, which makes it hard to investigate the effect of speaker variations in lip reading. Therefore, a large-scale lip reading database with speaker information, beyond the constrained databases \cite{cooke2006grid,zhao2009ouluvs}, is needed for the future research.

In this paper, we propose a speaker-adaptive training method for lip reading by introducing an additional speaker-specific input, namely user-dependent padding. The proposed user-dependent padding is for narrowing the data distribution gap between training speakers and the target test speaker. Distinct to the previous methods of using speaker-specific inputs in ASR that modified the extracted feature by introducing additional layers \cite{xue2014fastadaptation,abdel2013codeicassp,miao2015SATivector,abdel2013codecnn}, the proposed user-dependent padding participates in the visual feature extraction stage so that the personal lip appearances and movements can be jointly considered during the feature encoding. Moreover, it can interact with pre-trained weights without the necessity of additional network parameters or finetuning the network. This has the advantage of simple adaptation steps of directly adaptable from a pre-trained model, while the previous works \cite{xue2014fastadaptation, abdel2013codeicassp,miao2015SATivector,abdel2013codecnn} need to train a speech recognition model that attached an adaptation network.
Finally, the user-dependent padding is optimized for each target speaker, so it can achieve the optimal performance for each speaker with one pre-trained lip reading model.

Specifically, we replace the padding of convolution layers in the pre-trained lip reading model with the proposed user-dependent padding so that the additional speaker-specific input can interact with the learned convolution filter without modifying the architecture and weight parameters. By doing this, we can naturally achieve a strong regularization effect by maintaining the pre-trained weight, whereas previous works \cite{li2006l2regularized,yu2013kldregularized} tried with regularization loss to retain the learned model knowledge. Finally, to remedy the speaker information insufficient problem in large-scale lip reading databases, we label and provide speaker identity of a popular audio-visual dataset, LRW \cite{chung2016lrw}, obtained in the wild environment, and name LRW-ID to distinguish it from the original seen-speaker setting of LRW. The effectiveness of the proposed method is verified on GRID \cite{cooke2006grid} and the newly designed unseen-speaker lip reading scenario of LRW-ID.

The main contributions of the paper are as follows, 1) we propose a novel speaker-adaptive lip reading framework which utilizes user-dependent padding. User-dependent padding has a negligible number of parameters compared to that of the model and can improve the lip reading performance for each target speaker, adaptively. Moreover, it does not require any additional network and finetuning of the pre-trained model, 2) to the best of our knowledge, this is the first work to investigate the speaker-adaptive lip reading on a large-scale database obtained in the wild. To this end, we label the speaker information of a well-known large-scale audio-visual database, LRW, and build a new unseen-speaker lip reading setting named LRW-ID, and 3) compared to the previous speaker-adaptive and -independent speech recognition methods, we set new state-of-the-art performances and show the proposed method is close to practical usage.

\section{Related Work}
\subsection{Lip Reading}
Lip reading is a task of recognizing speech by watching lip movements only, which is regarded as one of the challenging problems. With the great development of deep learning, many research efforts have been made to improve the performance of lip reading \cite{kim2021lip, hong2021speech, mira2022end, mira2022svts}. In word-level lip reading, \cite{stafylakis2017resnetlstm} constructed an architecture consists of a 3D convolution layer and 2D ResNet \cite{he2016resnet} as a front-end and LSTM as a back-end. Some studies \cite{weng2019twostream, xiao2020deformation} proposed two-stream networks to better capture the lip movements by using the raw video and the optical flow. Recent work \cite{martinez2020mstcn} improved the temporal encoding with Multi-Scale Temporal Convolutional Network (MS-TCN). In sentence-level lip reading, \cite{assael2016lipnet} proposed an end-to-end model that trained with Connectionist Temporal Classification (CTC) \cite{graves2006ctc} loss. \cite {chung2017lrs2} developed lip reading based on Seq2Seq architecture \cite{sutskever2014sequence}. Further architectural improvement was made by \cite{afouras2018deep} using Transformer \cite{vaswani2017attention}. 
Some studies have focused on bringing audio modal knowledge into visual modality \cite{afouras2020asrisallyouneed, zhao2020hearing, ren2021learningfrommaster, kim2021mmbridge}. They successfully complemented the insufficient speech information of lip video with the rich audio knowledge. For example, \cite{kim2021cromm,kim2022distinguishing} proposed Visual-Audio Memory that can recall the audio features with just using the input video. Finally, \cite{ma2021lira,chung2016syncnet} proposed methods of pre-training the network in a self-supervised manner and showed promising results in lip reading.

Even with the successful development of the lip reading techniques, the speaker dependency of learned model has not been well studied. Since different speakers have different lip appearances and movements, applying a trained lip reading model to an unseen speaker can show degraded performance \cite{assael2016lipnet}. To effectively utilize the trained model without performance degradation, a method of speaker adaptation 
should be developed. In this paper, we investigate the speaker dependency of a pre-trained lip reading model and propose a speaker-adaptive lip reading method that can effectively adapt to an unseen speaker.

\subsection{Speaker Adaptation}
Speaker adaptation technique has been mainly developed in the area of audio-based Automatic Speech Recognition (ASR). \cite{liao2013trainpart1} examined finetuning the different parts of the model how affects the performance. However, the finetuning methods are easily suffer from the overfitting problem, especially with a small number of adaptation data. To handle this, \cite{yu2013kldregularized} tried to prevent the model from overfitting by regularizing the adaptation. Some works \cite{seide2011addlayer1, li2010addlayer2} tried to augments the speech recognition model with additional speaker-dependent layers. In \cite{swietojanski2014lhuc}, a speaker dependent vector is added to every hidden layer of the trained speech recognition model and adapted on the test speaker. In recent, using meta-learning \cite{klejch2019meta} and generation \cite{huang2020SAspeechsynthesis} based methods were proposed.

In other approaches, some works proposed providing additional speaker-specific inputs to the model for adapting the trained model to unseen speakers. \cite{miao2015SATivector} proposed to utilize i-vectors \cite{dehak2010ivector} extracted at the speaker level to suppress the speaker variance. With an adaptation network, the i-vectors are converted to speaker-specific shifts that will be added to the original acoustic features. In \cite{abdel2013codecnn,abdel2013codeicassp,xue2014fastadaptation}, they proposed speaker-specific inputs, named speaker code, which can be learned during the adaptation for each speaker. They have the advantage of adapting large-size models using only a few adaptation data. However, as they require additional layers to encode the speaker code, they need to train the additional layers using training data before performing the adaptation. Therefore, they need speaker labels for both training and adaptation data, where the training data contains many speakers compared to the adaptation data.

Compared to the research efforts in ASR, the speaker adaptation method has not been studied much in lip reading. Combination of MLLT \cite{gopinath1998mllt} and speaker adaptation \cite{anastasakos1997mllr} is applied to lip reading in \cite{almajai2016lipreadingmllt}. \cite{kandala2019SAlipreading} proposed to utilize i-vector in lip reading. These methods are evaluated with a constrained dataset with few speakers, due to the lack of speaker labels in public lip reading databases, and they need the speaker information of whole training data which is usually large.

In this paper, we develop a speaker-adaptive lip reading that utilizes user-dependent padding as speaker-specific inputs. Different from the previous methods, the proposed method does not need any additional network and the speaker information of the entire training data that utilized for pre-training. Instead, the proposed method can participate in the visual speech extraction stage of the visual front-end without modifying the network parameters, and just need the speaker information of adaptation data which is usually small (\eg, 1 minute). 

\section{Methods}
\begin{figure*}[t]
	\centering
	\centerline{\includegraphics[width=10.5cm]{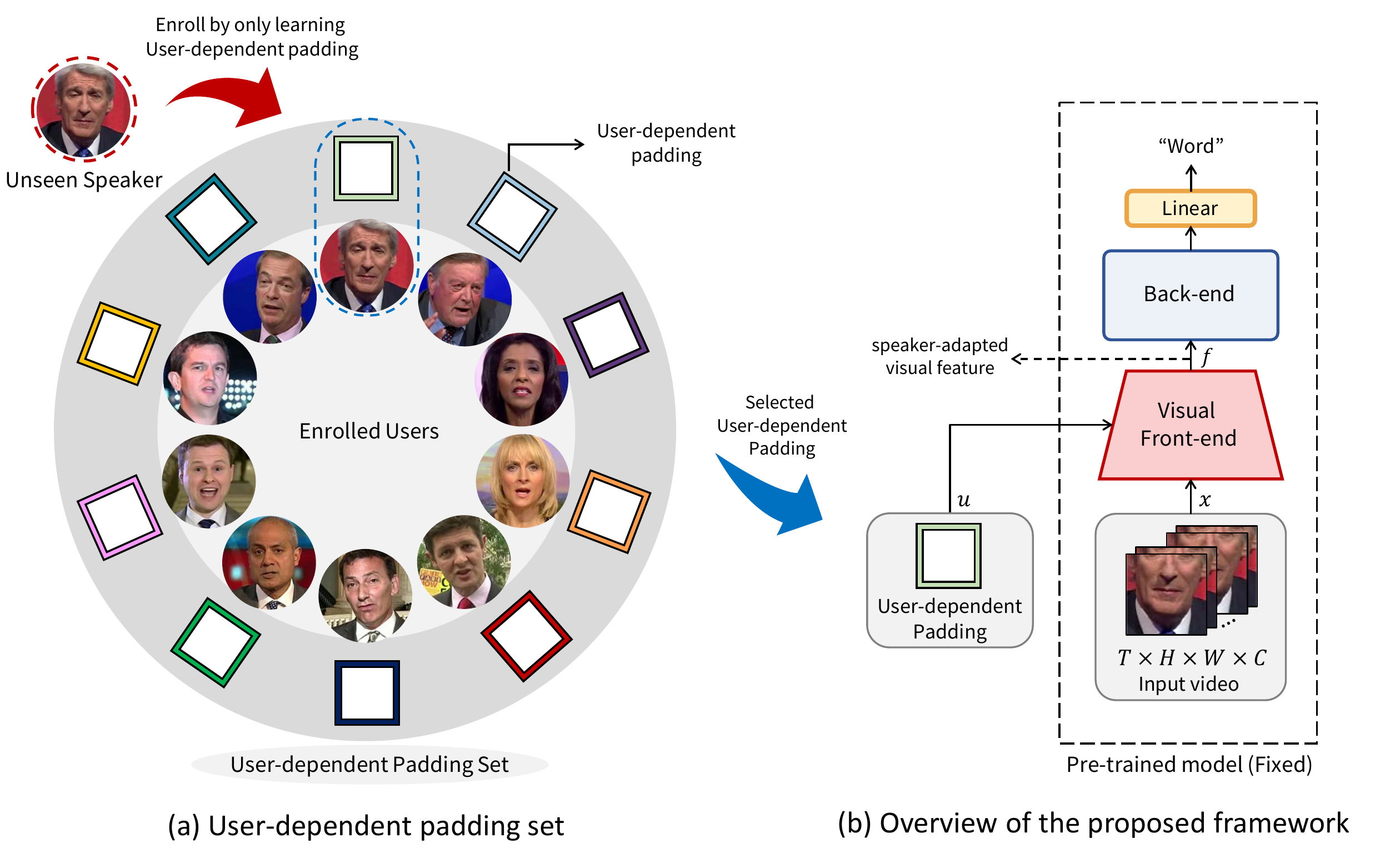}}
	\caption{Overview of the proposed framework. (a) When an unseen speaker is coming, the proposed framework can enroll the speaker by learning user-dependent padding only. (b) By using the user-dependent padding matched to the input speaker, lip reading model can adaptively encode the visual features and achieve improved performance}
	\label{fig:1}
\end{figure*}

Let $\mathcal{S}=\{(\mathbf{X}^s, \mathbf{Y}^s)\}=\{(x^s_1, y^s_1), \dots, (x^s_{N_s},y^s_{N_s})\}$ be a set of $N_s$ training samples where $x^s_i$ is the $i$-th lip video and $y^s_i$ is the corresponding ground-truth label, $\mathcal{A}_j=\{(\mathbf{X}^{a_j},\mathbf{Y}^{a_j})\}=\{(x^{a_j}_1, y^{a_j}_1), \dots, (x^{a_j}_{N_{a_j}}, y^{a_j}_{N_{a_j}})\}$ be a set of $N_{a_j}$ adaptation data of the $j$-th target speaker not appear in $\mathcal{S}$, and $\mathcal{T}_j=\{x^{t_j}_1,\dots,x^{t_j}_{N_{t_j}}\}$ be a test set of the target speaker.
With a pre-trained lip reading model learned on a large dataset $\mathcal{S}$ containing various speakers, our objective is to adapt the pre-trained model to the $j$-th unseen speaker using $\mathcal{A}_j$ containing a small number of data (\ie, $N_{a_j} \ll N_s$) in a supervised manner, thus achieving improved performance on the test data $\mathcal{T}_j$ of $j$-th unseen speaker. Otherwise, if the adaptation data $\mathcal{A}_j$ is not available, we try to adapt the pre-trained model directly on $\mathcal{T}_j$ in an unsupervised way. The overview of the proposed framework is shown in Fig. \ref{fig:1}.

A lip reading model is usually composed of a front-end $\mathcal{F}$ which extract visual features $f$ of lips, and back-end $\mathcal{B}$ which encodes the dynamics and predict the speech from the encoded visual features $f$. The training of a lip reading model can be achieved by updating the weight parameters $\theta$ of the front-end $\mathcal{F}$ and the back-end $\mathcal{B}$ through back-propagation of the loss computed using cross-entropy or CTC loss \cite{graves2006ctc} functions $\mathcal{L}(\cdot)$. It can be written as follows,
\begin{align}
    \label{eq:1}
    \theta^* = \argmin_\theta\mathcal{L}(\mathbf{Y}^s, \hat{\mathbf{Y}}), \quad \text{where} \,\, \hat{\mathbf{Y}} = (\mathcal{B}\circ \mathcal{F})_\theta(\mathbf{X}^s),
\end{align}
where $\theta^*$ is the parameters of the pre-trained lip reading model on a large dataset $\mathcal{S}$. With the pre-trained model, our goal is to encode speaker-adapted visual features $f$ according to the input speaker for improving performance.

\subsection{User-dependent Padding}
\begin{figure*}[t]
	\centering
	\centerline{\includegraphics[width=12.2cm]{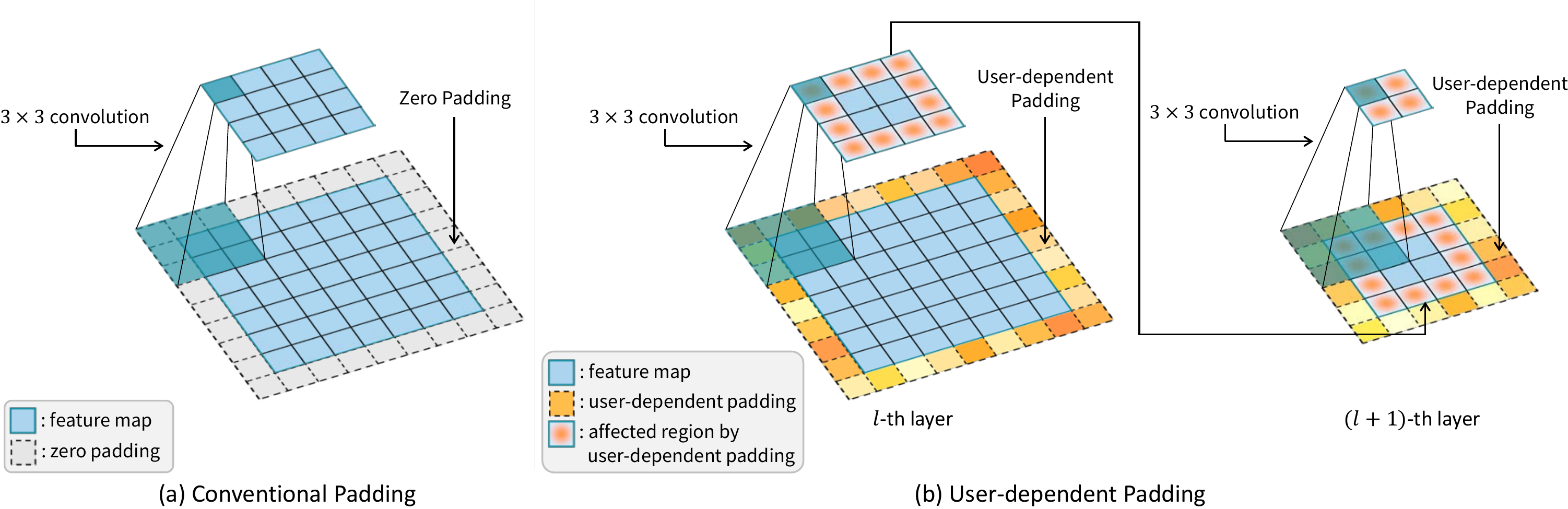}}
	\caption{Example of padded convolution with 3$\times$3 kernel and stride 2. (a) Conventional padding such as zero, reflect, and constant padding. (b) The proposed user-dependent padding can gradually affect entire visual features as it moves to deeper layers. 
	}
	\label{fig:2}
\end{figure*}
In ASR, to adapt the trained model on a new speaker, \cite{abdel2013codeicassp,miao2015SATivector,abdel2013codecnn,xue2014fastadaptation} proposed using a speaker code or i-vector \cite{dehak2010ivector} as an additional input to the trained model. The additional speaker-specific input is encoded with an additional network to modify the extracted acoustic features adaptively to the input speaker. However, they need to train the newly added network or fine-tune the entire network including the additional network on training data $\mathcal{S}$, before performing speaker adaptation.

We try to adapt the lip reading model on the $j$-th target speaker without introducing additional network and modifying the pre-trained weight $\theta^*$. Instead, we introduce an additional input $u$, called user-dependent padding, to the network which can interact with the pre-trained convolution filters in the front-end $\mathcal{F}$. Predictions of the lip reading model with the proposed additional inputs, user-dependent padding, can be written as, $\hat{\mathbf{Y}} = (\mathcal{B}\circ \mathcal{F})(\mathbf{X}^{t_j},u^{j}) = \mathcal{B}(\mathcal{F}(\mathbf{X}^{t_j},u^{j}))$. With the provided additional inputs, our desire is to allow the model to consider the personal lip characteristics during visual feature embedding. To make the additional input $u$ participate in the visual feature encoding without modifying the weight parameters, we utilize the region of padding in the CNN.

In CNN, for convolving the features with a kernel, padding is usually employed to maintain or control the output feature size. Conventionally used padding is zero padding, reflect padding, and constant padding. These padded region is also convolved with a learned kernel during the convolutions as shown in Fig. \ref{fig:2}a. We utilize these potential regions to insert the additional inputs. That is, the user-dependent padding is applied for the padding before convolution, instead of the conventional padding (\eg, zeros) used during pre-training, as shown in Fig. \ref{fig:2}b. If we assume that the pre-trained lip reading model is trained with zero padding, the optimization of a lip reading model in Eq. (\ref{eq:1}) can be re-written as,
\begin{align}
    \theta^* = \argmin_\theta\mathcal{L}(\mathbf{Y}^s, \hat{\mathbf{Y}}), \quad \text{where} \,\, \hat{\mathbf{Y}} = (\mathcal{B}\circ \mathcal{F})_\theta(\mathbf{X}^s, \mathbf{0}),
\end{align}
where $\mathbf{0}$ represents the zero inputs to be applied padding before the convolution operations. Now, we can provide additional inputs to the front-end by simply changing the zero inputs with the proposed user-dependent padding $u$, without modifying the learned weight parameters or using additional layers.

Then, with the proposed user-dependent padding $u$, the speaker-adaptation of a lip reading model on $j$-th unseen speaker using the adaptation data $\mathcal{A}_j$ can be achieved with the following equations,
\begin{align}
    u^{j*} = \argmin_{u^j}\mathcal{L}(\mathbf{Y}^{a_j}, \hat{\mathbf{Y}}), \quad \text{where} \,\, \hat{\mathbf{Y}} = (\mathcal{B}\circ \mathcal{F})_{\theta^*}(\mathbf{X}^{a_j}, u^j),
\end{align}
where $u^{j*}$ represents the learned user-dependent padding for the $j$-th speaker. Please note that we only optimize the user-dependent padding $u$, while maintaining the learned pre-trained knowledge $\theta^*$. Otherwise, if the adaptation data $\mathcal{A}$ is not available, the user-dependent padding also can be trained directly on $\mathcal{T}$ via any unsupervised training method such as self-training \cite{mei2020STUDA} and adversarial training \cite{tzeng2017adversarial} which are proven to be effective in unsupervised domain adaptation.

The user-dependent padding can affect the entire visual feature map as the layers go deeper, as shown in Fig. \ref{fig:2}b. Therefore, different from the previous methods that modify the extracted features \cite{xue2014fastadaptation,abdel2013codeicassp,miao2015SATivector}, the proposed user-dependent padding can participate in the whole visual feature encoding stages, so the personal lip appearances of the input speaker can be considered during the feature embedding. Finally, the user-dependent padding consumes small memory compared to the model, $\theta^*$, and this makes it possible to provide customized speech recognition services. All we need is one well-trained model in a central system and user-dependent paddings that can be deployed on personal mobile devices.

\subsection{LRW-ID}
\label{sec:3.2}
In order to develop and evaluate speaker-adaptive lip reading method, a dataset containing speaker information is essential. However, publicly available lip reading databases that contain speaker information are captured in a constrained environment and have a small number of speakers \cite{cooke2006grid,zhao2009ouluvs}, which is limited in evaluating the developed speaker-adaptive lip reading method. To remedy this problem, we clustered and labeled the speaker information of a large-scale unconstrained audio-visual dataset, LRW \cite{chung2016lrw}. Then, we split the train and test set without speaker overlapping which named as LRW-ID to distinguish it with the original splits of the dataset. Specifically, the speaker information of the LRW is labeled with the similar pipeline of \cite{anvari2019pipeline} as follows,

{\bf 1) Feature extraction.} 
In order to represent the speaker feature of a video, we employ a powerful face recognition system of ArcFace \cite{deng2019arcface}. We employ ResNet-101 \cite{he2016resnet} model pre-trained on MS-Celeb-1M \cite{guo2016msceleb1m}. From each video in LRW composed of 29 frames, 5 frames are randomly chosen for feature extraction. Face detection and alignment are performed using RetinaFace \cite{deng2020retinaface}. Then, the video-level speaker representation is obtained by averaging that of 5 frames embedded through the pre-trained face recognition model.

{\bf 2) Clustering.} With the obtained video-level speaker representations, speaker clustering is performed. For this stage, we perform face identification between video and clusters. Specifically, if the cosine similarity between a given video and all clusters is lower than a threshold $t_1$, a new cluster is introduced for the video. Otherwise, the video is assigned to a cluster showing the highest similarity. The speaker feature representing the cluster is updated with a new assigned video, with a momentum $m$ as, $C_k=\textit{norm}(m \times C_k + (1-m) \times f_l)$, where $C_k$ indicates face feature of cluster $k$, $f_l$ represents the normalized speaker feature of a video $l$ that assigned to cluster $k$, and $\textit{norm}(\cdot)$ represents l2 normalization.

{\bf 3) Face verification.} Due to the imperfection of clustering algorithms, having false positive samples are inevitable. To minimize the error, we should remove the false positive samples that different speakers are assigned to one cluster. To this end, face verification is performed between all samples in a cluster. Specifically, samples in the same cluster are compared by using their video-level speaker representations, and the cluster is split if they are detected as not the same person (\ie, the similarity is lower than a threshold $t_2$).

{\bf 4) Face identification.} In this stage, we deal with the multiple clusters of one speaker which should be merged. To handle this, face identification is performed between clusters. To represent the cluster-level speaker feature, the video-level speaker representations of all videos in the cluster are averaged. Each cluster is compared with the other clusters, and it is merged with multiple top-similarity clusters above a threshold $t_3$. 

{\bf 5) Manual correction.} Even if we merge the clusters through the previous step, we find that there still exist multiple clusters of the same speaker. Usually, they are not merged in the previous step due to the extreme differences in illumination and pose variations of faces that result in low similarities of face representations. To handle this, we extract the candidate clusters that exist in the boundary by using a lower threshold $t_4$ than used before, and manually inspect whether the clusters are from the same person or not.

The thresholds are empirically set by examining the quality of resulted clusters by humans as 0.41, 0.63, 0.63, and 0.59 for $t_1$, $t_2$, $t_3$, and $t_4$. The total number of labeled speakers through the above pipeline is 17,580 which is large compared to the previously used data \cite{cooke2006grid} for speaker adaptation. Therefore, it is very useful to evaluate the speaker-dependency of a lip reading model trained with large speaker variations. 
We choose 20 speakers who contain more than 900 videos to construct the test and adaptation (or validation) sets.
Information of the 20 selected speakers for the test is shown in Table \ref{table:2}. Since the classes that appear in adaptation and test sets are not perfectly overlapped, it is important that the speaker-adaptive method not be overfitted to the adaptation dataset.

\begin{table}[t]
    \renewcommand{\arraystretch}{1.2}
	\centering
	\caption{Selected 20 speakers for the test from LRW-ID. `\% Overlap class' represents how many word classes are overlapped between adaptation and test sets}
	\resizebox{0.97\linewidth}{!}{
	\begin{tabular}{ccccccccccc}
	\Xhline{3\arrayrulewidth}
	\makecell{\textbf{Speaker Number}\\\textbf{(Speaker ID)}} & \makecell{\textbf{S1} \\ (\#4243)} & \makecell{\textbf{S2} \\(\#5125)} & \makecell{\textbf{S3} \\(\#6003)} & \makecell{\textbf{S4}\\ (\#7184)} & \makecell{\textbf{S5} \\(\#9335)} & \makecell{\textbf{S6}\\ (\#9368)} & \makecell{\textbf{S7} \\(\#9438)} & \makecell{\textbf{S8} \\(\#9653)} & \makecell{\textbf{S9} \\(\#10209)} & \makecell{\textbf{S10}\\ (\#10293)} \\ \hline
    \# Tot. class   & 316 & 402 & 478 & 494 & 453 & 421 & 497 & 365 & 411 & 358 \\ \hline
    \# Tot. video  & 1130 & 1486 & 2381 & 6542 & 4116 & 1900 & 14478 & 1245 & 1490 & 1477 \\ \hline
    \makecell{\# Adapt. video\\(\# Word class)} & \makecell{565\\(252)} & \makecell{743\\(329)} & \makecell{1190\\(425)} & \makecell{3271\\(473)} & \makecell{2058\\(418)} & \makecell{950\\(346)} & \makecell{7239\\(493)} & \makecell{622\\(282)} & \makecell{745\\(316)} &\makecell{738\\(290)} \\ \hline
    \makecell{\# Test video\\(\# Word class)} & \makecell{565\\(240)} & \makecell{743\\(313)} & \makecell{1191\\(416)} & \makecell{3271\\(476)} & \makecell{2058\\(412)} & \makecell{950\\(346)} & \makecell{7239\\(495)} & \makecell{623\\(290)} & \makecell{745\\(330)} &\makecell{739\\(294)} \\ \hline
    \% Overlap class & 73.3 & 76.7 & 87.3 & 95.6 & 91.5 & 78.3 & 99.2 & 71.4 & 71.2 & 76.9 \\ 
    \Xhline{3\arrayrulewidth}
    \multicolumn{11}{c}{} \\ [-1.0ex]
    \Xhline{3\arrayrulewidth}
	\makecell{\textbf{Speaker Number} \\ \textbf{(Speaker ID)}} & \makecell{\textbf{S11} \\ (\#10587)} & \makecell{\textbf{S12} \\(\#11041)} & \makecell{\textbf{S13} \\(\#11777)} & \makecell{\textbf{S14}\\ (\#11875)} & \makecell{\textbf{S15} \\(\#11910)} & \makecell{\textbf{S16}\\ (\#13287)} & \makecell{\textbf{S17} \\(\#13786)} & \makecell{\textbf{S18} \\(\#15545)} & \makecell{\textbf{S19} \\(\#15769)} & \makecell{\textbf{S20}\\ (\#17378)} \\ \hline
    \# Tot. class   & 350 & 475 & 365 & 235 & 304 & 346 & 370 & 456 & 313 & 477 \\ \hline
    \# Tot. video  & 1106 & 5480 & 1743 & 2800 & 950 & 1213 & 1654 & 4126 & 936 & 3586 \\ \hline
    \makecell{\# Adapt. video\\(\# Word class)} & \makecell{553\\(258)} & \makecell{2740\\(455)} & \makecell{871\\(303)} & \makecell{1400\\(195)} & \makecell{476\\(236)} & \makecell{606\\(264)} & \makecell{827\\(298)} & \makecell{2063\\(419)} & \makecell{468\\(237)} &\makecell{1793\\(441)} \\ \hline
    \makecell{\# Test video\\(\# Word class)} & \makecell{553\\(268)} & \makecell{2740\\(447)} & \makecell{872\\(311)} & \makecell{1400\\(191)} & \makecell{476\\(239)} & \makecell{607\\(263)} & \makecell{827\\(303)} & \makecell{2063\\(426)} & \makecell{468\\(231)} &\makecell{1793\\(424)} \\ \hline
    \% Overlap class & 65.7 & 95.5 & 80.1 & 79.1 & 71.5 & 68.8 & 76.2 & 91.3 & 67.1 & 91.5 \\
    \Xhline{3\arrayrulewidth}
	\end{tabular} 	\label{table:2}
	}
\end{table}

\section{Experiments}
We evaluate the effectiveness of the proposed user-dependent padding on both sentence- and word-level lip reading databases. Moreover, we conduct experiments in two different adaptation settings, supervised adaptation where a small amount of adaptation data (\eg, under 5 minutes) is required and unsupervised adaptation where no supervision is required for the speaker adaptation.

\subsection{Dataset}
{\bf GRID} corpus \cite{cooke2006grid} is a popular sentence-level lip reading dataset. It is composed of sentences following the fixed grammar from 34 speakers. Videos are 3 seconds long, thus every 20 videos compose 1 minute. We follow the unseen speaker split of \cite{assael2016lipnet} that speakers 1, 2, 20, and 22 are used for test and the remainder is used for training. For the supervised adaptation setting, we split half of the data (\ie, about 500 videos) from each test speaker to construct the candidate dataset for adaptation and the others for the test data. For the unsupervised adaptation setting, all data from the test speakers are utilized for the test. For the performance measurement, Word Error Rate (WER) in percentage is utilized.

{\bf LRW-ID} is a speaker labeled version of LRW\cite{chung2016lrw}, a word-level lip reading dataset, as described in Sec. \ref{sec:3.2}. Each video is 1.16 seconds, thus 52, 155, and 259 videos compose 1, 3, and 5 minutes. 
For the supervised adaptation, the adaptation set is used for the speaker adaptation. For the unsupervised adaptation, only the test set is used. Word accuracy (\%) is utilized for the metric. 

\subsection{Baselines and Implementation Details}
Videos are pre-processed following \cite{kim2021cromm}. For LRW-ID, videos are cropped into 136 $\times$ 136 centered at the lip, resized into 112 $\times$ 112, and converted into grayscale. For GRID, the lip region is cropped and resized into a size of 64 $\times$ 128.

{\bf Baseline lip reading model.}
For the sentence-level lip reading, we utilize a modified network of LipNet \cite{assael2016lipnet}, which consists of three 3D convolutions and two 2D convolutions for the front-end, and two layered bi-GRU for the back-end. It is trained with CTC loss function \cite{graves2006ctc} with word tokens, and beam search with beam width 100 is utilized for the decoding. For the word-level lip reading, we employ an architecture of \cite{martinez2020mstcn}, which consists of ResNet-18 \cite{he2016resnet} for the front-end and MS-TCN \cite{martinez2020mstcn} for the back-end, and train the model using cross-entropy loss. AdamW optimizer \cite{kingma2014adam,loshchilov2017adamw} with an initial learning rate of 0.001, and batch size of 112 and 220 are utilized, respectively on GRID and LRW-ID.

{\bf Speaker-invariant.}
We borrow a speaker-invariant ASR method \cite{meng2018sitASR} into lip reading, which trains the model via adversarial learning to suppress the speaker information from the encoded features, to compare the effectiveness of the proposed method with a speaker-invariant speech recognition model. Specifically, an additional speaker classifier is introduced which classifies the speaker identity from the encoded visual feature. The sign of gradient calculated from the speaker classifier is reversed before backpropagated through the front-end \cite{ganin2015grl}, thus the front-end learns to suppress the speaker information from the encoded visual feature, while the speaker classifier attempts to find the speaker information from the encoded visual feature in an adversarial manner.

\begin{table}[t]
\renewcommand{\arraystretch}{1.2}
\renewcommand{\tabcolsep}{2.1mm}
\begin{minipage}{.51\linewidth}
    \centering
    \caption{Adaptation result using different time lengths of adaptation data on GRID}
    \resizebox{1.0\linewidth}{!}{
        \begin{tabular}[b]{cccccc}
        \Xhline{3\arrayrulewidth}
        \textbf{Adapt. min} & \textbf{S1} & \textbf{S2} & \textbf{S20} & \textbf{S22} & \textbf{Mean} \\ \hline
        Baseline & 17.04 & 9.02 & 10.33 & 8.13 & 11.12 \\
        1min & 10.65 & 4.20 & 7.77 & 4.59 & 6.80 \\ 
        3min & 9.35 & 3.75 & 6.88 & 4.27 & 6.05 \\ 
        5min & 8.78 & 3.45 & 6.49 & 3.99 & 5.67 \\  \Xhline{3\arrayrulewidth}
        \end{tabular} \label{table:3}
    }
\end{minipage}\hfill
\begin{minipage}{.47\linewidth}
    \centering
    \caption{Ablation results by using different padding layers with different amounts of adaptation data on LRW-ID}
    \resizebox{1.0\linewidth}{!}{
    	\begin{tabular}{cccc}
    	\Xhline{3\arrayrulewidth}
    	\textbf{Adapt. min} & \textbf{5 layers} & \textbf{11 layers} & \textbf{17 layers} \\ \hline
        1min & 86.54 & 86.81 & 87.06\\ 
        3min & 86.69 & 87.12 & 87.61\\ 
        5min & 86.85 & 87.31 & 87.91\\  \Xhline{3\arrayrulewidth}
    	\end{tabular}  \label{table:7}
    	}
\end{minipage}
\end{table}

{\bf Speaker Code.} 
We bring a popular speaker-adaptation method \cite{abdel2013codeicassp} of ASR which utilizes speaker code as additional inputs with additional layers, to compare the effectiveness of the proposed method with a speaker-adaptive model. For GRID, we use 128, 64, and 32 dimensions of speaker code with three additional fully connected layers which correspond to Adaptation Network of \cite{abdel2013codeicassp}, to transform the visual feature encoded from the front-end. For LRW, 256, 128, and 64 dimensions of speaker code are utilized. The training procedures are as follows, 1) bring a pre-trained lip reading model, 2) only train Adaptation Network and the speaker code after attaching them to the pre-trained model using the training dataset $\mathcal{S}$ while other network parameters are fixed, and 3) perform adaptation by training speaker code only using the adaptation dataset $\mathcal{A}$.

{\bf User-dependent Padding.} 
We utilize all padded convolutions to insert the user-dependent padding. For GRID, user-dependent padding is inserted before every 5 convolutional layers, and 17 convolutional layers for LRW-ID. The user-dependent padding is initialized with the padding used during pre-training (\ie, zero) and updated with a learning rate of 0.01. As the proposed method does not need an additional adaptation network, the training procedures can be simple as follows, 1) bring a pre-trained lip reading model, and 2) perform adaptation by updating the user-dependent padding only using the adaptation dataset $\mathcal{A}$.

\subsection{Supervised Adaptation}
{\bf Adaptation results using data under 5 minute.} To investigate the effectiveness of the proposed user-dependent padding, we adapt the lip reading model by using a small number of adaptation data. Specifically, we utilize 1, 3, and 5 minutes length of videos for adaptation, which might be relatively easily obtained in a practical situation. For reliable experimental results, each experiment is performed in 5 folds with different adaptation samples and the mean performance is reported. The results on GRID are shown in Table \ref{table:3}. The baseline achieves 11.12\% mean WER on four unseen speakers. By using 1 minute of adaptation data, the performances are significantly improved in all speakers by achieving mean WER of 6.80\%. Specifically, the WER of speaker 1 (s1) is improved by about 6.4\% WER from the baseline. By using adaptation data of 3 minutes, the mean WER is further improved to 6.05\%. Finally, adapting on 5 minutes video achieves 5.67\% WER. The adaptation results of each speaker on LRW-ID are shown in Table \ref{table:4}. The baseline model achieves 85.85\% mean word accuracy and it is improved by 1.21\% by adapting the model with 1 minute of adaptation video. Using more adaptation data further improves the performance. The mean word accuracy achieves 87.61\% and 87.91\% with 3 and 5 minutes adaptation data, respectively. This shows the effectiveness of the speaker-adaptation in lip reading that even if the model is trained with various speakers over 17,000, we can still improve the performance for unseen speakers through the adaptation. 

\begin{table}[t]
    \renewcommand{\arraystretch}{1.2}
	\centering
	\caption{Adaptation result using different time lengths of adaptation data on LRW-ID}
	\resizebox{0.97\linewidth}{!}{
	\begin{tabular}{ccccccccccc}
	\Xhline{3\arrayrulewidth}
	\multicolumn{1}{c|}{\makecell{\textbf{Adapt.}\\\textbf{min}}}
	& \makecell{\textbf{S1} \\ (\#4243)} & \makecell{\textbf{S2} \\(\#5125)} & \makecell{\textbf{S3} \\(\#6003)} & \makecell{\textbf{S4}\\ (\#7184)} & \makecell{\textbf{S5} \\(\#9335)} & \makecell{\textbf{S6}\\ (\#9368)} & \makecell{\textbf{S7} \\(\#9438)} & \makecell{\textbf{S8} \\(\#9653)} & \makecell{\textbf{S9} \\(\#10209)} & \makecell{\textbf{S10}\\ (\#10293)} \\ \hline
    \multicolumn{1}{c|}{Baseline} & 75.93 & 80.08 & 84.13 & 89.36 & 77.70 & 84.53 & 91.12 & 77.05 & 88.46 & 81.33 \\ 
    \multicolumn{1}{c|}{1min} & 78.94 & 82.15 & 85.14 & 89.39 & 81.68 & 85.07 & 91.57 & 80.06 & 88.46 & 84.00  \\ 
    \multicolumn{1}{c|}{3min} & 80.00 & 82.26 & 85.74 & 89.43 & 82.93 & 85.75 & 92.00 & 81.38 & 88.70 & 85.20  \\ 
    \multicolumn{1}{c|}{5min} & 81.10 & 82.77 & 85.79 & 89.64 & 83.83 & 86.21 & 92.05 & 81.86 & 88.75 & 85.68  \\  \Xhline{3\arrayrulewidth}
    \multicolumn{11}{c}{} \\ [-1.0ex]
    \Xhline{3\arrayrulewidth}
	\makecell{\textbf{S11} \\ (\#10587)} & \makecell{\textbf{S12} \\(\#11041)} & \makecell{\textbf{S13} \\(\#11777)} & \makecell{\textbf{S14}\\ (\#11875)} & \makecell{\textbf{S15} \\(\#11910)} & \makecell{\textbf{S16}\\ (\#13287)} & \makecell{\textbf{S17} \\(\#13786)} & \makecell{\textbf{S18} \\(\#15545)} & \makecell{\textbf{S19} \\(\#15769)} & \makecell{\textbf{S20}\\ (\#17378)} & \multicolumn{1}{|c}{\textbf{Mean}}\\ \hline
    73.78 & 86.83 & 88.07 & 85.79 & 72.69 & 75.95 & 81.74 & 87.01 & 88.25 & 86.67 & \multicolumn{1}{|c}{85.85}\\ 
    79.96 & 87.07 & 88.14 & 90.60 & 74.83 & 77.33 & 82.01 & 87.30 & 89.87 & 87.52 & \multicolumn{1}{|c}{87.06}\\ 
    81.88 & 87.60 & 88.17 & 91.54 & 76.89 & 77.86 & 82.06 & 87.44 & 90.17 & 88.20 & \multicolumn{1}{|c}{87.61}\\ 
    82.10 & 88.04 & 88.56 & 91.76 & 78.19 & 78.48 & 82.44 & 87.42 & 89.74 & 88.66 & \multicolumn{1}{|c}{87.91}\\  \Xhline{3\arrayrulewidth}
	\end{tabular} 	\label{table:4}
	}
\end{table}

\hfill \break
{\bf Comparison with previous methods.} We compare the adaptation results of the proposed method with the previous methods in ASR described in Sec. 4.2. The mean WER and mean word accuracy are reported in Table \ref{table:5} and the best two performances are highlighted in bold. The speaker-invariant model \cite{meng2018sitASR} improves the performance on both GRID and LRW-ID by suppressing the speaker variations. The speaker-adaptive method \cite{abdel2013codeicassp} which utilizes speaker code for the additional speaker-specific input also shows improved performances when the adaptation is performed, except for the 1 minute adaptation on LRW-ID. 
Even if the 1 minute adaptation on LRW-ID dataset is very challenging due to the small number of adaptation data, the proposed method robustly enhances the lip reading performances regardless of the adaptation video lengths. 
Moreover, we also report the performance of using the proposed user-dependent padding onto the speaker-invariant model \cite{meng2018sitASR} (\ie, Proposed Method + SI). By jointly applying the speaker-invariant and -adaptive techniques, we can further improve the overall lip reading performance.

\begin{table}[t]
	\renewcommand{\arraystretch}{1.2}
	\renewcommand{\tabcolsep}{1.8mm}
    \centering
    \caption{Performance comparisons with speaker-invariant and -adaptive methods}
    \resizebox{0.75\linewidth}{!}{
	\begin{tabular}{ccccp{0.01cm}ccc}
	\Xhline{3\arrayrulewidth}
	\multirow{2}{*}{\textbf{Method}} & \multicolumn{3}{c}{\textbf{GRID (WER $\downarrow$)}} & & \multicolumn{3}{c}{\textbf{LRW-ID (ACC $\uparrow$)}}\\ \cline{2-4} \cline{6-8}
	& \textbf{1min} & \textbf{3min} & \textbf{5min} & & \textbf{1min} & \textbf{3min} & \textbf{5min} \\ \hline
	Baseline \cite{assael2016lipnet,martinez2020mstcn} & 11.12 & 11.12 & 11.12 & & 85.85 & 85.85 & 85.85 \\
	Speaker-invariant (SI) \cite{meng2018sitASR} & 10.60 & 10.60 & 10.60 & & 86.55 & 86.55 & 86.55 \\ 
	Speaker code \cite{abdel2013codeicassp} & \textbf{6.77} & 6.32 & 6.21 & & 85.50 & 86.31 & 86.99 \\ \hline
	\textbf{Proposed Method} & \textbf{6.80} & \textbf{6.05} & \textbf{5.67} & & \textbf{87.06} & \textbf{87.61} & \textbf{87.91} \\ 
	\textbf{Proposed Method + SI} & 6.85 & \textbf{6.00} & \textbf{5.80} & & \textbf{87.59} & \textbf{88.14} & \textbf{88.48} \\
	\Xhline{3\arrayrulewidth}
	\end{tabular}  \label{table:5}
	}
\end{table}

\hfill \break
{\bf Adaptation results using more data.} 
In this experiment, we investigate the effectiveness of the proposed method when more adaptation data is available. Since the number of available adaptation data may be different for each individual person in practice, we perform experiments using 10, 30, 50, 70, 100\% of the adaptation data of each speaker. The mean results of all test speakers on GRID are shown in the second row of Table \ref{table:6}. Training the user-dependent padding using more adaptation data further improves the lip reading performances. When we utilize 100\% of the adaptation dataset (about 25 minutes), the model achieves 4.65\% WER which is improved over 6.4\% WER from the baseline. The last row of Table \ref{table:6} shows the mean results of all test speakers on LRW-ID, and it shows consistent results with the sentence-level lip reading by achieving steadily improved performances. 

\begin{table}[t]
    \renewcommand{\arraystretch}{1.2}
	\renewcommand{\tabcolsep}{2.1mm}
	\centering
	\caption{Adaptation result by using different rate of adaptation data}
	\resizebox{0.65\linewidth}{!}{
	\begin{tabular}{ccccccc}
	\Xhline{3\arrayrulewidth}
	\textbf{Adapt. \%} & \textbf{\,0\%} & \textbf{10\%} & \textbf{30\%} & \textbf{50\%} & \textbf{70\%} & \textbf{100\%} \\ \hline
    \textbf{GRID (WER $\downarrow$)} & 11.12 & 6.05 & 5.15 & 5.02 & 4.86 & 4.65 \\ 
    \textbf{LRW-ID (ACC $\uparrow$)} & 85.85 & 87.35 & 88.08 & 88.52 & 88.74 & 88.92 \\ 
    \Xhline{3\arrayrulewidth}
	\end{tabular} \label{table:6}
	}
\end{table}

\begin{figure}[t]
\begin{minipage}[c]{0.49\linewidth}
    \centering
    \includegraphics[width=1.0\linewidth]{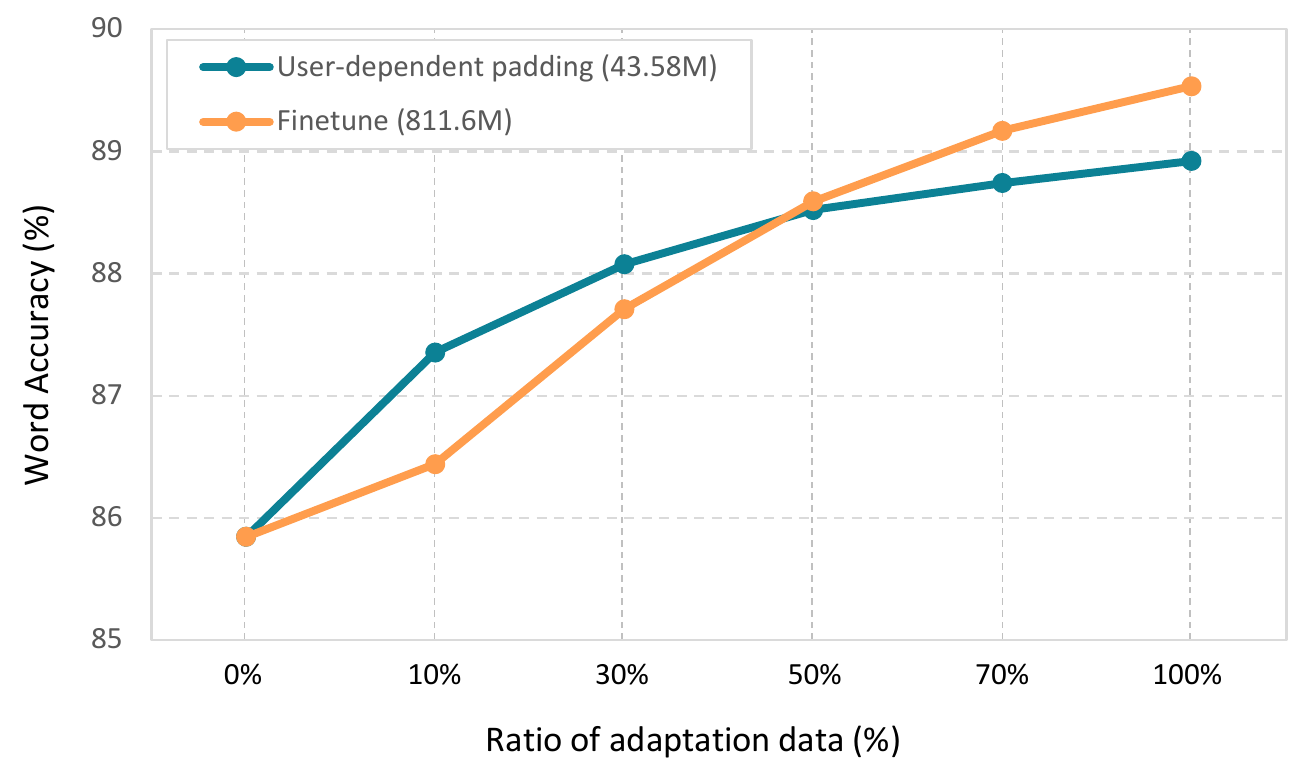} 
    \captionof{figure}{Performance comparisons with finetuning method on LRW-ID}
    \label{fig:3}
\end{minipage}
\hfill
\begin{minipage}[c]{0.47\linewidth}
\renewcommand{\arraystretch}{1.2}
\renewcommand{\tabcolsep}{2.1mm}
    \centering
    \captionof{table}{Unsupervised adaptation results on GRID and LRW-ID}
    \resizebox{1.0\linewidth}{!}{
    	\begin{tabular}{ccc}
    	\Xhline{3\arrayrulewidth}
    	\textbf{Method} & \makecell{\textbf{GRID} \\ \textbf{(WER $\downarrow$)}} & \makecell{\textbf{LRW-ID} \\ \textbf{(ACC $\uparrow$)}}\\ \hline
    	MS-TCN \cite{martinez2020mstcn} & - & 85.85 \\
    	CroMM-VSR \cite{kim2021cromm} & - & 87.30 \\
    	LipNet \cite{assael2016lipnet} & 11.12 & - \\ 
    	TVSR-Net \cite{yang2020SIlipreading} & 9.1 & - \\
    	DVSR-Net \cite{zhang2021SIlipreading} & 7.8 & - \\
    	Visual i-vector \cite{kandala2019SAlipreading} & 7.3 & - \\ \hline
    	\textbf{Proposed Method} & \textbf{7.2} & \textbf{87.51} \\ \Xhline{3\arrayrulewidth}
    	\end{tabular} \label{table:8}
    	}
\end{minipage}
\end{figure}

\hfill \break
{\bf Comparison with finetuning.} 
We compare the effectiveness of the user-dependent padding with finetuning. To this end, the entire model parameters are finetuned from pre-trained lip reading model on the adaptation data of LRW-ID. This yields the total number of 20 speaker-specific lip reading models which results in a total of $20\times40.58\text{M}=811.6\text{M}$ parameters. Fig. \ref{fig:3} shows the comparison results on LRW-ID dataset. When a small amount of adaptation data is utilized (\ie, less than 30\%), the user-dependent padding surpasses the finetuning. This is because finetuning the entire model parameter with a small number of data can be easily overfitted to the classes that appear in adaptation data. On the other hand, the user-dependent padding largely improves the performance with just 10\% of adaptation data which shows the significance of the proposed method in the small data setting. When more than 50\% of the adaptation data are utilized, the finetuning shows better performance than the user-dependent padding. Please note that user-dependent paddings for 20 speakers have $20\times0.15\text{M}=3\text{M}$ parameters, thus we just require $3\text{M}+40.58\text{M}=43.58\text{M}$ parameters including that of one pre-trained model, which is about 19 times smaller than using user-specific lip reading models (\ie, 811.6M). This result shows the user-dependent padding is effective and practical even if enough adaptation data is available.

\hfill \break
{\bf Ablation Study.} 
Finally, we investigate the effect of the number of padding layers for user-dependent padding. To this end, we vary the number of layers for inserting the user-dependent padding from the total 17 layers of ResNet-18. We use 5, 11, and 17 layers and 1, 3, and 5 minutes of adaptation data for the experiments. Table \ref{table:7} shows the ablation results by using different padding layers on LRW-ID. When 1 minute of adaptation data is utilized, the performances are less varying by the different number of layers, and when more adaptation data is used, the performance gain by using more layers becomes larger, while that of using 5 layers is marginal. This means that we can use a small number of layers for the user-dependent padding to avoid overfitting when we have a very small amount of adaptation data, and as the adaptation data increases, we can increase the padding layers accordingly to achieve high performance.

\subsection{Unsupervised Adaptation}
One advantage of the user-dependent padding is that it does not depend on specific training methods. That means we can bring any unsupervised learning method to adapt our lip reading model on an unseen speaker when adaptation dataset $\mathcal{A}$ is not available. To verify this, we employ a self-training method \cite{yarowsky1995ST1,lee2013ST2,xie2020ST3,vesely2013ST4} that pseudo labels the unlabeled samples with the pre-trained lip reading model before adaptation. We utilize one of the simplest form of self-training that chooses the pseudo labels by inspecting the model confidences. Specifically, we use the model predictions having approximated beam confidence larger than 0.9 for GRID and the model predictions with over 0.8 confidence for LRW-ID, as the pseudo labels to perform adaptation. Table \ref{table:8} shows the mean performance of test speakers in the unsupervised adaptation setting on GRID and LRW-ID, and the comparisons with the state-of-the-art methods. The results for each speaker can be found in the supplemental material. The proposed method sets new state-of-the-art performances on both word- and sentence-level lip reading without using adaptation data $\mathcal{A}$, and this shows the effectiveness and the practicality of the proposed user-dependent padding in speaker adaptation. 

\section{Conclusion}
In this paper, we propose a speaker adaptation method for lip reading, named user-dependent padding. The proposed user-dependent padding can cooperate with the pre-trained model without modifying the architecture and learned weight parameters. The effectiveness of the proposed method is verified on both sentence- and word-level lip reading. Through the experiment, we show that with just a few amount of adaptation data, the lip reading performance for unseen speakers can be further improved, even if the model is pre-trained with many utterances from thousands of speakers. Finally, for the future research on speaker-adaptation in lip reading, we label speaker of a popular lip reading database, LRW, and build a new unseen-speaker scenario named LRW-ID.

\hfill \break
{\bf Acknowledgment.} This work was supported by Institute of Information \& communications Technology Planning \& Evaluation (IITP) grant funded by the Korea government(MSIT) (No.2022-0-00124, Development of Artificial Intelligence Technology for Self-Improving Competency-Aware Learning Capabilities)

\clearpage
%
%
\bibliographystyle{splncs04}
\bibliography{egbib}

\begin{thebibliography}{10}
\providecommand{\url}[1]{\texttt{#1}}
\providecommand{\urlprefix}{URL }
\providecommand{\doi}[1]{https://doi.org/#1}

\bibitem{abdel2013codeicassp}
Abdel-Hamid, O., Jiang, H.: Fast speaker adaptation of hybrid nn/hmm model for
  speech recognition based on discriminative learning of speaker code. In: 2013
  IEEE International Conference on Acoustics, Speech and Signal Processing. pp.
  7942--7946. IEEE (2013)

\bibitem{abdel2013codecnn}
Abdel-Hamid, O., Jiang, H.: Rapid and effective speaker adaptation of
  convolutional neural network based models for speech recognition. In:
  INTERSPEECH. pp. 1248--1252 (2013)

\bibitem{afouras2018deep}
Afouras, T., Chung, J.S., Senior, A., Vinyals, O., Zisserman, A.: Deep
  audio-visual speech recognition. IEEE transactions on pattern analysis and
  machine intelligence  (2018)

\bibitem{afouras2018lrs3}
Afouras, T., Chung, J.S., Zisserman, A.: Lrs3-ted: a large-scale dataset for
  visual speech recognition. arXiv preprint arXiv:1809.00496  (2018)

\bibitem{afouras2020asrisallyouneed}
Afouras, T., Chung, J.S., Zisserman, A.: Asr is all you need: Cross-modal
  distillation for lip reading. In: ICASSP 2020-2020 IEEE International
  Conference on Acoustics, Speech and Signal Processing (ICASSP). pp.
  2143--2147. IEEE (2020)

\bibitem{almajai2016lipreadingmllt}
Almajai, I., Cox, S., Harvey, R., Lan, Y.: Improved speaker independent lip
  reading using speaker adaptive training and deep neural networks. In: 2016
  IEEE International Conference on Acoustics, Speech and Signal Processing
  (ICASSP). pp. 2722--2726. IEEE (2016)

\bibitem{anastasakos1997mllr}
Anastasakos, T., McDonough, J., Makhoul, J.: Speaker adaptive training: A
  maximum likelihood approach to speaker normalization. In: 1997 IEEE
  International Conference on Acoustics, Speech, and Signal Processing. vol.~2,
  pp. 1043--1046. IEEE (1997)

\bibitem{anvari2019pipeline}
Anvari, Z., Athitsos, V.: A pipeline for automated face dataset creation from
  unlabeled images. In: Proceedings of the 12th ACM International Conference on
  PErvasive Technologies Related to Assistive Environments. pp. 227--235 (2019)

\bibitem{assael2016lipnet}
Assael, Y.M., Shillingford, B., Whiteson, S., De~Freitas, N.: Lipnet:
  End-to-end sentence-level lipreading. arXiv preprint arXiv:1611.01599  (2016)

\bibitem{chung2017lrs2}
Chung, J.S., Senior, A., Vinyals, O., Zisserman, A.: Lip reading sentences in
  the wild. In: 2017 IEEE conference on computer vision and pattern recognition
  (CVPR). pp. 3444--3453. IEEE (2017)

\bibitem{chung2016lrw}
Chung, J.S., Zisserman, A.: Lip reading in the wild. In: Asian conference on
  computer vision. pp. 87--103. Springer (2016)

\bibitem{chung2016syncnet}
Chung, J.S., Zisserman, A.: Out of time: automated lip sync in the wild. In:
  Asian conference on computer vision. pp. 251--263. Springer (2016)

\bibitem{cooke2006grid}
Cooke, M., Barker, J., Cunningham, S., Shao, X.: An audio-visual corpus for
  speech perception and automatic speech recognition. The Journal of the
  Acoustical Society of America  \textbf{120}(5),  2421--2424 (2006)

\bibitem{dehak2010ivector}
Dehak, N., Kenny, P.J., Dehak, R., Dumouchel, P., Ouellet, P.: Front-end factor
  analysis for speaker verification. IEEE Transactions on Audio, Speech, and
  Language Processing  \textbf{19}(4),  788--798 (2010)

\bibitem{deng2020retinaface}
Deng, J., Guo, J., Ververas, E., Kotsia, I., Zafeiriou, S.: Retinaface:
  Single-shot multi-level face localisation in the wild. In: Proceedings of the
  IEEE/CVF Conference on Computer Vision and Pattern Recognition. pp.
  5203--5212 (2020)

\bibitem{deng2019arcface}
Deng, J., Guo, J., Xue, N., Zafeiriou, S.: Arcface: Additive angular margin
  loss for deep face recognition. In: Proceedings of the IEEE/CVF conference on
  computer vision and pattern recognition. pp. 4690--4699 (2019)

\bibitem{digalakis1995unseen}
Digalakis, V.V., Rtischev, D., Neumeyer, L.G.: Speaker adaptation using
  constrained estimation of gaussian mixtures. IEEE Transactions on speech and
  Audio Processing  \textbf{3}(5),  357--366 (1995)

\bibitem{ganin2015grl}
Ganin, Y., Lempitsky, V.: Unsupervised domain adaptation by backpropagation.
  In: International conference on machine learning. pp. 1180--1189. PMLR (2015)

\bibitem{gopinath1998mllt}
Gopinath, R.A.: Maximum likelihood modeling with gaussian distributions for
  classification. In: Proceedings of the 1998 IEEE International Conference on
  Acoustics, Speech and Signal Processing, ICASSP'98 (Cat. No. 98CH36181).
  vol.~2, pp. 661--664. IEEE (1998)

\bibitem{graves2006ctc}
Graves, A., Fern{\'a}ndez, S., Gomez, F., Schmidhuber, J.: Connectionist
  temporal classification: labelling unsegmented sequence data with recurrent
  neural networks. In: Proceedings of the 23rd international conference on
  Machine learning. pp. 369--376 (2006)

\bibitem{guo2016msceleb1m}
Guo, Y., Zhang, L., Hu, Y., He, X., Gao, J.: Ms-celeb-1m: A dataset and
  benchmark for large-scale face recognition. In: European conference on
  computer vision. pp. 87--102. Springer (2016)

\bibitem{he2016resnet}
He, K., Zhang, X., Ren, S., Sun, J.: Deep residual learning for image
  recognition. In: Proceedings of the IEEE conference on computer vision and
  pattern recognition. pp. 770--778 (2016)

\bibitem{hong2021speech}
Hong, J., Kim, M., Park, S.J., Ro, Y.M.: Speech reconstruction with reminiscent
  sound via visual voice memory. IEEE/ACM Transactions on Audio, Speech, and
  Language Processing  \textbf{29},  3654--3667 (2021)

\bibitem{huang2020SAspeechsynthesis}
Huang, Y., He, L., Wei, W., Gale, W., Li, J., Gong, Y.: Using personalized
  speech synthesis and neural language generator for rapid speaker adaptation.
  In: ICASSP 2020-2020 IEEE International Conference on Acoustics, Speech and
  Signal Processing (ICASSP). pp. 7399--7403. IEEE (2020)

\bibitem{kandala2019SAlipreading}
Kandala, P.A., Thanda, A., Margam, D.K., Aralikatti, R.C., Sharma, T., Roy, S.,
  Venkatesan, S.M.: Speaker adaptation for lip-reading using visual identity
  vectors. In: INTERSPEECH. pp. 2758--2762 (2019)

\bibitem{kim2021cromm}
Kim, M., Hong, J., Park, S.J., Ro, Y.M.: Cromm-vsr: Cross-modal memory
  augmented visual speech recognition. IEEE Transactions on Multimedia  (2021)

\bibitem{kim2021mmbridge}
Kim, M., Hong, J., Park, S.J., Ro, Y.M.: Multi-modality associative bridging
  through memory: Speech sound recollected from face video. In: Proceedings of
  the IEEE/CVF International Conference on Computer Vision. pp. 296--306 (2021)

\bibitem{kim2021lip}
Kim, M., Hong, J., Ro, Y.M.: Lip to speech synthesis with visual context
  attentional gan. Advances in Neural Information Processing Systems
  \textbf{34},  2758--2770 (2021)

\bibitem{kim2022distinguishing}
Kim, M., Yeo, J.H., Ro, Y.M.: Distinguishing homophenes using multi-head
  visual-audio memory for lip reading. In: Proceedings of the 36th AAAI
  Conference on Artificial Intelligence, Vancouver, BC, Canada. vol.~22 (2022)

\bibitem{kingma2014adam}
Kingma, D.P., Ba, J.: Adam: A method for stochastic optimization. arXiv
  preprint arXiv:1412.6980  (2014)

\bibitem{klejch2019meta}
Klejch, O., Fainberg, J., Bell, P., Renals, S.: Speaker adaptive training using
  model agnostic meta-learning. In: 2019 IEEE Automatic Speech Recognition and
  Understanding Workshop (ASRU). pp. 881--888. IEEE (2019)

\bibitem{lee2013ST2}
Lee, D.H., et~al.: Pseudo-label: The simple and efficient semi-supervised
  learning method for deep neural networks. In: Workshop on challenges in
  representation learning, ICML. vol.~3, p.~896 (2013)

\bibitem{li2010addlayer2}
Li, B., Sim, K.C.: Comparison of discriminative input and output
  transformations for speaker adaptation in the hybrid nn/hmm systems. In:
  Eleventh Annual Conference of the International Speech Communication
  Association (2010)

\bibitem{li2006l2regularized}
Li, X., Bilmes, J.: Regularized adaptation of discriminative classifiers. In:
  2006 IEEE International Conference on Acoustics Speech and Signal Processing
  Proceedings. vol.~1, pp.~I--I. IEEE (2006)

\bibitem{liao2013trainpart1}
Liao, H., McDermott, E., Senior, A.: Large scale deep neural network acoustic
  modeling with semi-supervised training data for youtube video transcription.
  In: 2013 IEEE Workshop on Automatic Speech Recognition and Understanding. pp.
  368--373. IEEE (2013)

\bibitem{loshchilov2017adamw}
Loshchilov, I., Hutter, F.: Decoupled weight decay regularization. arXiv
  preprint arXiv:1711.05101  (2017)

\bibitem{ma2021lira}
Ma, P., Mira, R., Petridis, S., Schuller, B.W., Pantic, M.: Lira: Learning
  visual speech representations from audio through self-supervision. arXiv
  preprint arXiv:2106.09171  (2021)

\bibitem{martinez2020mstcn}
Martinez, B., Ma, P., Petridis, S., Pantic, M.: Lipreading using temporal
  convolutional networks. In: ICASSP 2020-2020 IEEE International Conference on
  Acoustics, Speech and Signal Processing (ICASSP). pp. 6319--6323. IEEE (2020)

\bibitem{mei2020STUDA}
Mei, K., Zhu, C., Zou, J., Zhang, S.: Instance adaptive self-training for
  unsupervised domain adaptation. In: European conference on computer vision.
  pp. 415--430. Springer (2020)

\bibitem{meng2018sitASR}
Meng, Z., Li, J., Chen, Z., Zhao, Y., Mazalov, V., Gong, Y., Juang, B.H.:
  Speaker-invariant training via adversarial learning. In: 2018 IEEE
  International Conference on Acoustics, Speech and Signal Processing (ICASSP).
  pp. 5969--5973. IEEE (2018)

\bibitem{miao2014towardsSAT}
Miao, Y., Zhang, H., Metze, F.: Towards speaker adaptive training of deep
  neural network acoustic models. In: Fifteenth annual conference of the
  international speech communication association (2014)

\bibitem{miao2015SATivector}
Miao, Y., Zhang, H., Metze, F.: Speaker adaptive training of deep neural
  network acoustic models using i-vectors. IEEE/ACM Transactions on Audio,
  Speech, and Language Processing  \textbf{23}(11),  1938--1949 (2015)

\bibitem{mira2022svts}
Mira, R., Haliassos, A., Petridis, S., Schuller, B.W., Pantic, M.: Svts:
  Scalable video-to-speech synthesis. arXiv preprint arXiv:2205.02058  (2022)

\bibitem{mira2022end}
Mira, R., Vougioukas, K., Ma, P., Petridis, S., Schuller, B.W., Pantic, M.:
  End-to-end video-to-speech synthesis using generative adversarial networks.
  IEEE Transactions on Cybernetics  (2022)

\bibitem{neto1995lin}
Neto, J., Almeida, L., Hochberg, M., Martins, C., Nunes, L., Renals, S.,
  Robinson, T.: Speaker-adaptation for hybrid hmm-ann continuous speech
  recognition system  (1995)

\bibitem{noda2014lipreadingconv}
Noda, K., Yamaguchi, Y., Nakadai, K., Okuno, H.G., Ogata, T.: Lipreading using
  convolutional neural network. In: fifteenth annual conference of the
  international speech communication association (2014)

\bibitem{petridis2018end}
Petridis, S., Stafylakis, T., Ma, P., Cai, F., Tzimiropoulos, G., Pantic, M.:
  End-to-end audiovisual speech recognition. In: 2018 IEEE international
  conference on acoustics, speech and signal processing (ICASSP). pp.
  6548--6552. IEEE (2018)

\bibitem{ren2021learningfrommaster}
Ren, S., Du, Y., Lv, J., Han, G., He, S.: Learning from the master: Distilling
  cross-modal advanced knowledge for lip reading. In: Proceedings of the
  IEEE/CVF Conference on Computer Vision and Pattern Recognition. pp.
  13325--13333 (2021)

\bibitem{seide2011addlayer1}
Seide, F., Li, G., Chen, X., Yu, D.: Feature engineering in context-dependent
  deep neural networks for conversational speech transcription. In: 2011 IEEE
  Workshop on Automatic Speech Recognition \& Understanding. pp. 24--29. IEEE
  (2011)

\bibitem{stafylakis2017resnetlstm}
Stafylakis, T., Tzimiropoulos, G.: Combining residual networks with lstms for
  lipreading. arXiv preprint arXiv:1703.04105  (2017)

\bibitem{sutskever2014sequence}
Sutskever, I., Vinyals, O., Le, Q.V.: Sequence to sequence learning with neural
  networks. Advances in neural information processing systems  \textbf{27}
  (2014)

\bibitem{swietojanski2014lhuc}
Swietojanski, P., Renals, S.: Learning hidden unit contributions for
  unsupervised speaker adaptation of neural network acoustic models. In: 2014
  IEEE Spoken Language Technology Workshop (SLT). pp. 171--176. IEEE (2014)

\bibitem{tzeng2017adversarial}
Tzeng, E., Hoffman, J., Saenko, K., Darrell, T.: Adversarial discriminative
  domain adaptation. In: Proceedings of the IEEE conference on computer vision
  and pattern recognition. pp. 7167--7176 (2017)

\bibitem{vaswani2017attention}
Vaswani, A., Shazeer, N., Parmar, N., Uszkoreit, J., Jones, L., Gomez, A.N.,
  Kaiser, {\L}., Polosukhin, I.: Attention is all you need. Advances in neural
  information processing systems  \textbf{30} (2017)

\bibitem{vesely2013ST4}
Vesel{\`y}, K., Hannemann, M., Burget, L.: Semi-supervised training of deep
  neural networks. In: 2013 IEEE Workshop on Automatic Speech Recognition and
  Understanding. pp. 267--272. IEEE (2013)

\bibitem{weng2019twostream}
Weng, X., Kitani, K.: Learning spatio-temporal features with two-stream deep 3d
  cnns for lipreading. arXiv preprint arXiv:1905.02540  (2019)

\bibitem{xiao2020deformation}
Xiao, J., Yang, S., Zhang, Y., Shan, S., Chen, X.: Deformation flow based
  two-stream network for lip reading. In: 2020 15th IEEE International
  Conference on Automatic Face and Gesture Recognition (FG 2020). pp. 364--370.
  IEEE (2020)

\bibitem{xie2020ST3}
Xie, Q., Luong, M.T., Hovy, E., Le, Q.V.: Self-training with noisy student
  improves imagenet classification. In: Proceedings of the IEEE/CVF conference
  on computer vision and pattern recognition. pp. 10687--10698 (2020)

\bibitem{xue2014fastadaptation}
Xue, S., Abdel-Hamid, O., Jiang, H., Dai, L., Liu, Q.: Fast adaptation of deep
  neural network based on discriminant codes for speech recognition. IEEE/ACM
  Transactions on Audio, Speech, and Language Processing  \textbf{22}(12),
  1713--1725 (2014)

\bibitem{yang2020SIlipreading}
Yang, C., Wang, S., Zhang, X., Zhu, Y.: Speaker-independent lipreading with
  limited data. In: 2020 IEEE International Conference on Image Processing
  (ICIP). pp. 2181--2185. IEEE (2020)

\bibitem{yang2019lrw1000}
Yang, S., Zhang, Y., Feng, D., Yang, M., Wang, C., Xiao, J., Long, K., Shan,
  S., Chen, X.: Lrw-1000: A naturally-distributed large-scale benchmark for lip
  reading in the wild. In: 2019 14th IEEE International Conference on Automatic
  Face \& Gesture Recognition (FG 2019). pp.~1--8. IEEE (2019)

\bibitem{yarowsky1995ST1}
Yarowsky, D.: Unsupervised word sense disambiguation rivaling supervised
  methods. In: 33rd annual meeting of the association for computational
  linguistics. pp. 189--196 (1995)

\bibitem{yu2013kldregularized}
Yu, D., Yao, K., Su, H., Li, G., Seide, F.: Kl-divergence regularized deep
  neural network adaptation for improved large vocabulary speech recognition.
  In: 2013 IEEE International Conference on Acoustics, Speech and Signal
  Processing. pp. 7893--7897. IEEE (2013)

\bibitem{zhang2021SIlipreading}
Zhang, Q., Wang, S., Chen, G.: Speaker-independent lipreading by disentangled
  representation learning. In: 2021 IEEE International Conference on Image
  Processing (ICIP). pp. 2493--2497. IEEE (2021)

\bibitem{zhao2009ouluvs}
Zhao, G., Barnard, M., Pietikainen, M.: Lipreading with local spatiotemporal
  descriptors. IEEE Transactions on Multimedia  \textbf{11}(7),  1254--1265
  (2009)

\bibitem{zhao2020hearing}
Zhao, Y., Xu, R., Wang, X., Hou, P., Tang, H., Song, M.: Hearing lips:
  Improving lip reading by distilling speech recognizers. In: Proceedings of
  the AAAI Conference on Artificial Intelligence. vol.~34, pp. 6917--6924
  (2020)

\end{thebibliography}
\end{document}